\documentclass{article}

\usepackage[dblblindworkshop,final]{neurips_2026}
\workshoptitle{JUDGe: Can We Trust the Judge?}

\usepackage[utf8]{inputenc}
\usepackage[T1]{fontenc}
\usepackage[hidelinks]{hyperref}
\usepackage{url}
\usepackage{booktabs}
\usepackage{amsfonts}
\usepackage{nicefrac}
\usepackage{microtype}
\usepackage{xcolor}
\usepackage{amsmath}
\usepackage{graphicx}
\usepackage{array}
\usepackage{placeins}

\title{Behind Aggregate Scores: What Systematically Varying Arena-Hard's LLM Judge Reveals}

\author{%
  Louise Yorke \\
  University College London \\
  London, UK \\
  \texttt{l.yorke@ucl.ac.uk}
}

\begin{document}

\maketitle
\begin{abstract}
LLM judges are used to determine which of two answers is better. Trusting these verdicts requires knowing how repeatable they are and how much they depend on the judge’s instructions. We show that similar aggregate scores can conceal substantial changes in individual verdicts, both when the same setup is repeated and when the instructions change. Using 500 Arena-Hard v2.0 prompts with two models’ answers held fixed, we run GPT-4.1 as the judge under all eight combinations of criteria, output format and task framing, and repeat every setup unchanged at temperature zero on a pinned model version. Rerunning the default setup moves the aggregate score by about one point but changes the win/tie/loss outcome on 21.2\% of prompts. Instruction changes introduce further sensitivity: in one setup, which asks for the verdict without reasoning and also drops several other instructions, outcomes change on 28.2 percentage points more prompts than the two setups’ unchanged reruns do on average, and the aggregate score falls by 4.09 points. In another, where the criteria prioritise accuracy, outcomes change on 6.7 percentage points more prompts by the same comparison, while the aggregate score barely moves. A stable aggregate score therefore does not establish that individual verdicts are repeatable. This distinction matters when those verdicts are used as preference labels or to assess individual responses. Evaluations that rely on individual verdicts should repeat each setup they use and report both aggregate scores and prompt-level agreement.
\end{abstract}

\section{Introduction}

Large language models are widely used to rank other language models. Pairwise win-rate benchmarks built on LLM judges inform model selection and alignment work \citep{zheng2023judging,li2024crowdsourced,dubois2024length}. Their judges depend on choices about criteria, output format, task description, judge model, answer order and verdict scoring. Many of these choices are defensible, but they are not always fully reported.

Aggregate scores rank systems on leaderboards \citep{chiang2024chatbot,li2024crowdsourced}, while individual prompt outcomes can supply preference labels, regression tests and safety-review evidence. A stable aggregate does not establish that these individual judgments are stable. We show that substantial changes in prompt-level outcomes can occur even when the aggregate barely moves.

Previous research has examined judge biases, including position, length and formatting effects \citep{wang2024large,dubois2024length,li2024crowdsourced}. Recent work also documents variation between unchanged runs at temperature zero \citep{wei2024systematic,norman2026reliability,ayyagari2026pinned} and between judge designs \citep{vaugrante2026design,zhang2026sensitive}. We examine both on the same prompts, comparing each instruction change with the test--retest disagreement of the two setups involved, that is, the share of prompts whose outcome changes when the same setup is simply run again. Our design draws on multi-analyst studies outside machine learning \citep[see Appendix~\ref{app:related}]{silberzahn2018many,botvinik2020variability,steegen2016increasing}.

We hold all 500 Arena-Hard v2.0 prompts and both models' responses fixed. GPT-4.1 judges them under eight setups crossing three choices: Arena-Hard's criteria or accuracy-prioritising criteria; reasoning before the verdict or the verdict label alone; and no task description or a coding-assistance description. We repeat every setup unchanged on a pinned model version. We also substitute Claude Sonnet 4.5 and Gemini 2.5 Flash under the default instructions. These substitutions were run only in May and are compared with the default setup's test--retest disagreement alone.

Rerunning the default setup moves the aggregate by about one point but changes 21.2\% of prompt outcomes. Compared with the default, the label-only setup changes outcomes on 28.2 percentage points more prompts than the two setups' unchanged reruns do on average. The corresponding excess is 6.7 points for accuracy-prioritising criteria and 3.8 points for coding-assistance framing. Across 17 runs, comprising two per setup and a third default run, only 131 of 496 complete prompts (26.4\%) retain the same outcome throughout. These comparisons concern instruction packages, not isolated wording changes.

Our contributions are to distinguish aggregate from prompt-level stability and to assess instruction sensitivity against the test--retest disagreement of the setups being compared. We recommend testing a small, prespecified set of alternative setups, repeating each unchanged, and reporting changes in aggregate scores and individual outcomes alongside each setup's test--retest disagreement.\footnote{Code, raw judge outputs for every run, and the analysis producing the main results are available at \url{https://github.com/Yorkel/llm-judge-release}.}

\section{Method}

\subsection{Data and judge setups}

We use the 500 prompts and fixed model responses from the Arena-Hard v2.0 hard split. GPT-4.1 mini is the candidate and o3-mini is the baseline model. Arena-Hard judges each pair twice, reversing response order. We recode the five verdict labels from $-2$ to $2$ so that positive values always favour the candidate, then average the two orders. Prompts and responses never change; only the judge setup varies.

The main experiment uses GPT-4.1 under eight instruction setups in a complete $2 \times 2 \times 2$ factorial. GPT-4.1 is the default judge in the Arena-Hard v2.0 code used here.\footnote{At commit \texttt{196f6b8}, the Arena-Hard documentation identifies Gemini 2.5 with style control as the official configuration and offers GPT-4.1 as an alternative for users of the OpenAI API.} First, we change the judging criteria. Arena-Hard's original package asks the judge to generate its own answer and consider helpfulness, relevance, concision, creativity, errors and missing information. Our alternative makes factual and technical accuracy primary: technical claims, code, calculations and reasoning must be accurate even when another response is more helpful. Helpfulness and safety remain constraints, and the judge does not generate a reference answer.

Second, we change the required output. The judge either reasons before returning a verdict or returns the verdict label alone. Under the original criteria, the label-only package also removes the reference-answer and error-checking instructions, so this is a package contrast rather than an isolated effect of visible reasoning.

Third, we change the task framing. The judge receives either no scope statement or a coding-assistance description. That description is accurate for the 253 coding prompts but not the 247 mathematics prompts, so we report both full-sample and coding-only comparisons. Appendix~\ref{app:prompts} gives the full instruction packages.

We also hold Arena-Hard's original instructions fixed and substitute Claude Sonnet 4.5 or Gemini 2.5 Flash for GPT-4.1. We test these two judges separately, as an illustrative comparison outside the controlled October design. Because provider code and generation settings also differ, these comparisons include differences in implementation as well as judge model.

The main results come from October 2026. We ran all eight GPT-4.1 setups twice and the default setup a third time, using the pinned version \texttt{gpt-4.1-2025-04-14} at temperature zero. All 17 runs were completed on the same day. Of 17,000 calls, 16,996 returned a parsed verdict (see Appendix~\ref{app:parsing} for the parsing rule). Prompts and responses were fixed throughout (Arena-Hard code commit \texttt{196f6b8}; dataset revision \texttt{15f3746}). Paired comparisons include only prompts with both response orders complete in every run being compared, leaving 496 or 497 prompts for comparisons with the default setup. Appendix~\ref{app:conditions} lists all setups and their counts. We use these repeats to measure each setup's test--retest disagreement. Temperature zero does not make hosted inference deterministic \citep{atil2024nondeterminism,he2025defeating}.

The Claude and Gemini comparisons come from an earlier experiment in May 2026, when each judge setup was run once. The GPT-4.1 default was repeated in August using the same unpinned model name. These comparisons therefore use only the default setup's test--retest disagreement, which may also include model changes between May and August. We report them separately from the October instruction comparisons.

\subsection{Outcomes and analysis}

Our primary aggregate outcome is the raw win-equivalent score. It maps the order-averaged verdicts onto a 0-to-100 scale, treating the five ordinal levels as equally spaced. This equal spacing is our choice: Arena-Hard's own scoring counts a strong preference as three wins, a slight preference as one and a tie as half a win. Scoring the comparisons in Table~\ref{tab:main} that way gives the same conclusions. Aggregate comparisons in Table~\ref{tab:main} use paired changes on the same complete prompts rather than differences between separately calculated condition means.

Our primary prompt-level outcome is a boundary change: a prompt moves between candidate win, baseline win or tie between two runs, whether the judge setup is unchanged or different. A prompt's outcome in a run comes from its order-averaged score: positive scores indicate a candidate win, negative scores a baseline win and zero a tie. Thus, $+0.5$ counts as a candidate win and $-0.5$ as a baseline win. Appendix~\ref{app:robustness} reports results with both counted as ties.

The sensitivity of a prompt is the number of the other seven instruction setups that change its outcome relative to the default setup. We relate sensitivity to the default setup's margin, taken from a run not used to count changes. Agreement between judge models is summarised with linearly weighted Cohen's $\kappa$, computed after re-binning the order-averaged scores to five ordinal tiers.

For each contrast we also report the excess over test--retest disagreement: the boundary-change rate against the default setup, averaged over every pairing of their runs, minus the average of the two setups' own test--retest disagreement, each the boundary-change rate between runs of the same setup. If runs are independent, this difference has expectation zero when the two setups give each prompt the same distribution of outcomes, including how often it changes on a rerun; a setup that is simply more consistent therefore also shows an excess, which we count as part of the effect of changing the instructions. Measured instead against the default setup's test--retest disagreement alone, the higher of the two in every comparison here, the excess is $+20.5$ points (95\% CI 16.4 to 24.5) for the label-only setup, $+5.7$ (3.1 to 8.5) for accuracy-prioritising criteria and $+2.1$ ($-0.2$ to $+4.5$) for the coding-assistance description; measured against the label-only setup's own, it is $+36.0$, and the reported $+28.2$ is the average of the two. For the earlier judge-model comparisons, we subtract only the default setup's May--August disagreement; Appendix~\ref{app:definitions} gives both estimators.

Unless stated otherwise, intervals are 95\% percentile bootstrap intervals from 20,000 draws that resample whole prompts with replacement, drawing all of a prompt's runs together so that paired quantities share their sampling variation. As an implementation check, scoring the May run of the default setup exactly as Arena-Hard does, with its weighted wins and style control fitted jointly with Arena-Hard's published judgments of 27 other models, gives GPT-4.1 mini 48.0\% (90\% interval 46.3\% to 50.3\%), against Arena-Hard's published 48.6\%. This check shows that the aggregate scores are similar; it does not show that the individual verdicts match.

\section{Results}

\subsection{Similar aggregate scores can conceal different prompt outcomes}

Across the eight GPT-4.1 instruction setups, raw win-equivalent scores on the paired samples used in Table~\ref{tab:main}, averaged over each setup's runs, range from 49.36 to 53.87. Similar scores come from different prompt-level outcomes. Among the 496 prompts with complete outcomes in every run, only 131 (26.4\%) remain a candidate win, baseline win or tie in every run of every setup. The other 365 prompts (73.6\%) change outcome in at least one run. Six prompts are tied in every run.

Table~\ref{tab:main} compares each GPT-4.1 setup with the default setup on the same complete prompts. The unchanged repeat is the first result and the reference for the rest. The default setup's three runs produce aggregate scores about one point apart (52.82, 54.02 and 53.50), yet 21.2\% of prompts (95\% CI 18.4 to 24.1) change their winner-or-tie outcome, averaged across the three pairwise comparisons. Each setup has its own test--retest disagreement: 17.7\% to 21.2\% for the setups that reason first and 2.0\% to 7.0\% for those that return the label alone. Table~\ref{tab:main} reports instruction comparisons both as raw boundary-change rates and as excesses over test--retest disagreement.

\begin{table}[!htbp]
\centering
\caption{Paired changes from the default setup in the October runs (\texttt{gpt-4.1-2025-04-14}). Scores are on a 0 to 100 scale; a different winner means a move between candidate win, baseline win and tie. Each setup's own test--retest disagreement comes from its two runs, the default setup's from its three. Excess subtracts the average of the default setup's test--retest disagreement and the setup's own. Intervals are 95\% percentile bootstrap intervals over prompts (20,000 draws).}
\label{tab:main}
\includegraphics[width=\linewidth]{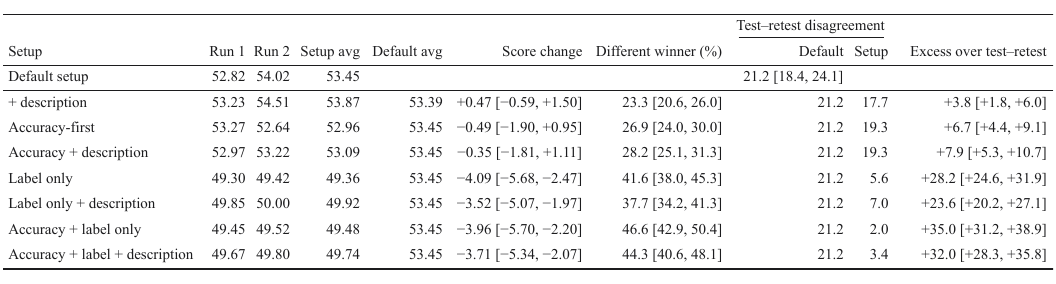}
\end{table}

The criteria package changes 26.9\% of prompts with little aggregate change ($-0.49$ points), 6.7 points above test--retest disagreement. The coding-assistance description adds a small 3.8 points (95\% CI 1.8 to 6.0), which is not distinguishable from zero against the default setup's test--retest disagreement alone. This excess is concentrated in the mathematics prompts, for which the description is inaccurate (Appendix~\ref{app:robustness}). Appendix~\ref{app:figures} visualises the aggregate and prompt-level changes and reports direct reversals.

In the May runs, which were not repeated in October, replacing GPT-4.1 with Gemini barely changed the aggregate but changed many prompt outcomes: the paired aggregate changed by 0.18 points while 38.3\% of prompts changed outcome, 19.0 points above the default setup's May--August disagreement; Claude lowered the aggregate by 3.97 points and changed 35.0\%. Agreement between GPT-4.1 and Gemini is moderate rather than close, with linearly weighted Cohen's $\kappa$ of 0.435 (95\% CI 0.378 to 0.492; 1,000 bootstrap draws); $\kappa$ is 0.549 between GPT-4.1 and Claude, and 0.478 between Claude and Gemini. The default setup's runs agree with each other at $\kappa = 0.763$ (95\% CI 0.730 to 0.794). Agreement between different judges was therefore lower than agreement between unchanged GPT-4.1 runs.

The label-only setup changes both the aggregate score and individual outcomes: it lowers the paired aggregate by 4.09 points and changes 41.6\% of prompt outcomes, 28.2 points above test--retest disagreement.\footnote{Averaged over the four pairs of setups that differ in only one choice, label-only output lowers the score by 3.72 points (95\% CI 2.38 to 5.03); the accuracy-first criteria ($-0.30$) and the coding description ($+0.36$) change it little.} Most of this change is movement into ties: 31.4\% of prompts become ties, 6.2\% leave a tie and 4.0\% reverse, as the tie rate rises from 18.5\% to 43.5\% (Table~\ref{tab:conditions}). The label-only judge rarely swaps winners but calls far more ties, which matters wherever ties are dropped or fall on the wrong side of a pass mark. Aggregate movement therefore does not reliably indicate how many underlying verdicts changed.

\subsection{Sensitivity concentrates near the default setup's decision boundary}

Prompt-level changes are most common when the default setup expresses only a weak preference. To measure the margin separately from the runs used to count changes, each prompt's margin is taken from a third run of the default setup. Under the criteria contrast, the mean absolute margin is 0.69 among changed prompts and 1.05 among unchanged prompts (two-sided Mann--Whitney $p = 3.6 \times 10^{-9}$). A prompt close to zero requires less movement to cross between a win and a tie or between a tie and a loss. The rate of change falls as the margin grows, and change rates are similar at the two highest margins. Appendix~\ref{app:figures} shows the full sensitivity distribution.

Changes also concentrate near ties when the default setup is repeated unchanged. Rerunning the default setup unchanged changes 43.4\% of prompts with a margin of zero, 33.8\% at a margin of 0.5, and 17.6\%, 4.3\% and 5.5\% at the three higher margins. Near-tied prompts change more often even when the instructions are unchanged. The ratio between the ends is roughly eight to one, so instability is not spread evenly over the benchmark: it is concentrated on a subset of prompts and largely absent elsewhere.

\begin{figure}[htbp]
\centering
\includegraphics[width=0.66\linewidth]{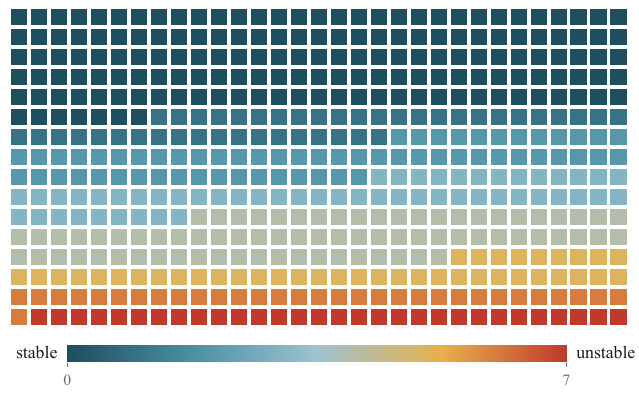}
\caption{Prompt-level stability across instruction setups. Each square is one prompt, shaded by how many of the seven other GPT-4.1 instruction setups change its winner-or-tie outcome relative to the default setup, comparing first runs, and ordered by that count. The 496 prompts with a verdict in every run are shown. In all, 162 prompts are unchanged under all seven setups and 30 change under all seven.}
\label{fig:stability}
\end{figure}

Of the 496 prompts with a verdict in every run, 162 (32.7\%) keep the default setup's outcome under all seven other instruction setups and 30 (6.0\%) change under all seven, comparing each setup's first run with the default setup's first run. Figure~\ref{fig:stability} shows the distribution. The stable group has a mean absolute margin of 1.30 against 0.65 for the always-changing group (Mann--Whitney $p = 5.7 \times 10^{-6}$). The margin was measured in a separate run from those used to identify these groups.

Coding-only, missing-output and style-adjustment checks are reported in Appendix~\ref{app:robustness}; none changes the interpretation of the raw paired comparisons.

\section{Discussion}

Aggregate stability does not establish prompt-level stability, and the unchanged repeat shows this most simply. Running one setup twice moves the headline score by about a point but changes a fifth of the individual verdicts. Subtracting test--retest disagreement helps assess instruction changes, but does not make individual verdicts more repeatable. The two levels behave differently because prompts move in both directions in similar numbers. Of the 105 prompts that change outcome between the default setup's first two runs, 56 move toward the candidate and 49 toward the baseline. These movements mostly cancel in the average, while the individual verdicts still change.

Changing the judge adds to this. Two of the changes we tested barely move the headline score but change 5.7 and 19.0 percentage points more prompts than a repeat of the default setup does: the accuracy-prioritising criteria, and Gemini in the May runs. Looking only at the aggregate would miss both the test--retest disagreement and the excess.

Some of the movement comes from how the two answer orders are combined. Each prompt is judged in both orders and the two five-point verdicts are averaged, which produces half-point values. If the judge calls a tie in one order and a slight preference in the other, the prompt scores $\pm 0.5$, and the sign rule records a win although the judge called a tie in one of the two orders. These half-point scores are over-represented among prompts that change: 39.0\% of changes between the default setup and the other instruction setups involve a score of $\pm 0.5$, against a base rate of 29.5\%. Counting $\pm 0.5$ as a tie instead lowers the default setup's test--retest disagreement from 21.2\% to 19.3\% and leaves every excess above zero (Appendix~\ref{app:robustness}). The sign rule therefore accounts for only a small part of the disagreement between runs.

Bias checks examine whether judges systematically favour particular positions, lengths or formats. In these repeats, changes towards the two models were similar in number. A judge could pass every usual bias check and still change a fifth of its verdicts on a second run. Repeating the run allows this variation to be measured.

Whether this matters depends on how the evaluation is used. Ranking two systems on a leaderboard depends on the aggregate, and for that the accuracy-prioritising criteria and the Gemini substitution give almost the same comparison. Position on a larger leaderboard is more fragile: scored with Arena-Hard's own method against its published judgments of 27 other models, the three unchanged October runs of the default setup place GPT-4.1 mini 11th, 7th and 11th of 28, because several neighbouring models sit within about two points of each other.
Individual verdicts matter when they become preference labels, flag responses for safety review, signal regressions or feed into release decisions, and for those uses the same judge changes alter many of the outcomes. A judge setup can therefore be reliable enough to rank two models but not to label individual prompts. About a quarter of the prompts keep the same outcome in all 17 GPT-4.1 runs, although this does not establish that those outcomes are correct.

We did not test whether combining judges improves reliability or accuracy. We compare agreement between different judges with agreement between unchanged runs of the same judge. Agreement across the three pairs of unchanged default runs averages $\kappa = 0.763$, so the 0.435 to 0.549 between judges should be compared with that figure, not with perfect agreement. In measurement terms this is a test--retest reliability, which psychology and education report before a test is used to decide about an individual \citep{aera2014standards}. We do not claim that 0.763 is an upper limit, since a different judge or a combined procedure could agree more closely with either run. Judge pipelines typically make one call per item and report no agreement statistic, so the variation in each call goes into the published score instead of being averaged out.

We recommend reporting repeatability alongside sensitivity to judge setup. Evaluations that rely on individual verdicts should choose a small set of plausible alternative judge setups in advance, run them alongside the main setup, and report three things: which setups were tested; the paired aggregate change and prompt-level change rate for each; and the same two figures for an unchanged repeat of each setup. Test--retest disagreement differs by setup, from 2.0\% to 21.2\% of prompts here, so the main setup's repeat cannot stand in for the others. Repeats show how much outcomes change without changing the instructions. The average test--retest disagreement of the setups compared is about four fifths of the raw change rate under coding-assistance framing and three quarters under accuracy-prioritising criteria.

These comparisons do not show which verdict is correct, or that one judge setup is better than another.

\section{Limitations}

Each instruction contrast changes a whole package of instructions. The output-format contrast is especially broad: under Arena-Hard's criteria, the label-only version removes several instructions as well as the requirement to explain the verdict. We therefore attribute results to the packages tested, not to individual instructions.

Each GPT-4.1 setup was run two or three times on one day, so each setup's test--retest disagreement rests on one to three pairs of runs and may vary between days. The excess over both setups' test--retest disagreement also assumes that runs are independent. The repeats measure the variation but do not explain it: we cannot tell whether the prompts that move are genuinely ambiguous or the judge is simply unstable on them, since both would produce the same pattern. We also have no human or ground-truth labels, apart from four prompts whose answers could be checked (Appendix~\ref{app:robustness}), so the study measures how much verdicts change, not which setup is more accurate. The Claude and Gemini conditions were run only in May, so their excess uses only the default setup's May--August disagreement. They also used different provider code, generation settings and, for Claude, a lower output-token limit, so they test the whole judge implementation, not just the model. Gemini ran with its internal reasoning budget set to zero, which matters because the output-format contrast is itself about reasoning. An incomplete third substitution, Gemini 2.5 Pro, is reported in Appendix~\ref{app:robustness}.

The results cover one benchmark, one model pair and one main judge model, GPT-4.1, with Claude and Gemini tested only once each. The model pair was chosen because its published comparison is close to 50\%, which makes movement across the decision boundary easier to see. The change rates should not be taken as typical for model pairs with larger capability gaps. Response style was examined only in the May runs, through relative length and markdown density (Appendix~\ref{app:robustness}), so we do not establish how style contributes to the observed changes. The threshold-based classification in Appendix~\ref{app:thresholds} is descriptive, and the share of unanimous prompts does not depend on either threshold. Further work should test a second model pair with a wider capability gap, a second benchmark, repeats on different days, and whether majority verdicts over repeated runs reduce the instability.

\section{Conclusion}

With prompts and responses held fixed, an unchanged rerun moves the aggregate score by about a point but changes 21.2\% of prompt outcomes. The label-only setup changes 28.2 percentage points more prompts than test--retest disagreement and lowers the aggregate by 4.09 points; the accuracy-prioritising criteria add 6.7 points with little aggregate change, and coding-assistance framing adds 3.8 points. Evaluations that rely on individual verdicts should therefore report both levels and compare each change of setup with an unchanged repeat of each setup.

\bibliographystyle{plainnat}
\bibliography{custom}

\appendix

\section{Related work}
\label{app:related}

LLM-as-judge protocols emerged as a scalable substitute for human pairwise ratings \citep{zheng2023judging} and now sit inside many major open evaluation pipelines \citep{dubois2024length,li2024crowdsourced,lin2025wildbench}. Industry uses the same template for regression testing, safety auditing and capability benchmarking \citep{bai2022constitutional,inan2023llama,liang2022holistic}, and surveys document its spread \citep{li2025generation}. The literature has identified specific failure modes, including position bias \citep{wang2024large}, length bias \citep{dubois2024length,saito2023verbosity} and preferences for familiar formatting \citep{li2024crowdsourced}. Our analysis includes two common mitigations: answer-order swapping in every condition, and a style adjustment built on Arena-Hard's feature extractor. These approaches address bias within a fixed judge configuration. They do not test how verdicts move when the configuration itself changes.

That question has an established form outside machine learning. Seventy independent teams analysing one fMRI dataset reached systematically divergent findings \citep{botvinik2020variability}, and twenty-nine teams analysing one dataset in econometrics reached opposing conclusions \citep{silberzahn2018many}. The methodological response is to report the distribution of conclusions across the defensible analytic space rather than one path through it \citep{steegen2016increasing,simonsohn2020specification}. Within machine learning, underspecification \citep{d2022underspecification} and predictive multiplicity \citep{marx2020predictive,semenova2022existence} show that many models with similar measured performance can exist for one task. Here, we hold the model responses fixed and vary the judge. The measurement-validity tradition supplies the relevant distinction between a construct and its operationalisation \citep{jacobs2021measurement}.

The closest precursor outside judging is \citet{sclar2024quantifying}, which shows that prompt-format perturbations can change benchmark accuracy by tens of points. Related work extends format sensitivity to paraphrases and retrieval evaluation \citep{mizrahi2024state,salinas2024butterfly,arabzadeh2025human}. Two differences matter here. First, that work perturbs the prompt given to the system under test, whereas we perturb the instrument that judges it. Second, it reports aggregate swings, whereas our central result is that the aggregate can remain almost unchanged while many individual judgments change. A separate line of work aims to make a fixed judge more reliable through checklists \citep{lee2025checkeval}, trained evaluators \citep{kim2024prometheus} or uncertainty estimates \citep{sheng2025analyzing}. We address a different quantity: whether a reported result survives defensible changes to the judge.

Run-to-run variation in LLM judges has been measured directly. \citet{wei2024systematic} find that GPT-4o's pairwise verdicts are not fully self-consistent even at temperature zero; \citet{haldar2025rating} report low agreement between repeated runs of open-weight judges with sampling; \citet{schroeder2024trust} find no variation for locally run open-weight judges at temperature zero but substantial variation above it; and \citet{norman2026reliability} report test--retest reliability at temperature zero for a large set of judges, GPT-4.1 among them. Concurrently with this work, \citet{ayyagari2026pinned} reruns four judges ten times at temperature zero and finds that 0.13\% to 9.7\% of verdicts change, concentrated on close calls, while the principal judge's aggregate ranking holds. Our default setup's rate of 21.2\% is well above that range. Several features of our design could contribute: the model pair was chosen to sit near 50\%, so many prompts are close calls; each outcome combines two judge calls, one per answer order; and the sign rule turns a tie in one order and a slight lean in the other into a win, although counting $\pm 0.5$ as a tie lowers our rate only to 19.3\%. Output format may matter most: our label-only setups, at 2.0\% to 7.0\%, fall within their range. Separately, \citet{vaugrante2026design} and \citet{zhang2026sensitive} vary judge design and report how much the verdicts change, without a rerun to compare against. We measure both kinds of change on the same prompts and report each design change as an excess over test--retest disagreement.

\section{Additional figures}
\label{app:figures}

\begin{figure}[htbp]
\centering
\includegraphics[width=\linewidth]{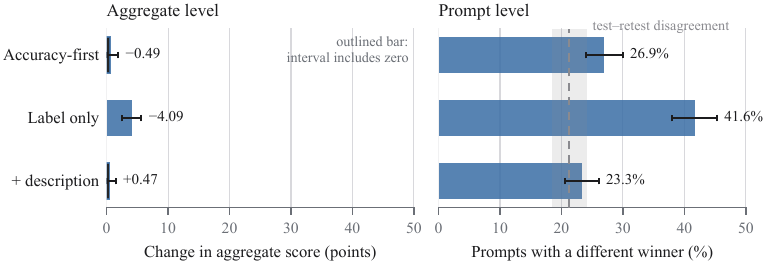}
\caption{Paired aggregate changes and prompt-level change rates for the three single-choice setups in the October runs. Both panels share a 0 to 50 point scale and the same row order; the left panel plots the magnitude of the aggregate change with the signed value annotated, and outlines rather than fills any bar whose interval includes zero. The dashed line and shaded band mark the default setup's own test--retest disagreement, the rate at which two of its runs give a different winner, with its 95\% interval. Setups can remain close at the aggregate level while differing substantially in prompt-level outcomes. Whether a setup exceeds test--retest disagreement is decided by the excess in Table~\ref{tab:main}, not by the bar's position relative to the line.}
\label{fig:aggregate-boundary}
\end{figure}

Most boundary changes are movement into or out of a tie rather than a direct reversal of which model wins. Direct reversals account for 5.1\% of paired prompts under the criteria package, 4.0\% under label-only output and 2.9\% under coding-assistance framing, and in the May runs 8.5\% under Claude and 7.9\% under Gemini. The default setup reverses 2.0\% on a rerun, and the label-only setups almost never reverse themselves (0.0\% to 0.4\%). Against the average reversal rate of the two setups' reruns, the criteria package ($+3.3$, CI $+2.0$ to $+4.6$) and the label-only setups ($+2.2$ to $+2.8$) reverse more often than rerunning does, while coding-assistance framing reverses no more often than rerunning does ($+0.9$, CI $-0.03$ to $+2.05$). In May, Claude ($+6.7$) and Gemini ($+6.2$) also reversed more often than the default setup's May--August repeat.

\begin{figure}[htbp]
\centering
\includegraphics[width=\linewidth]{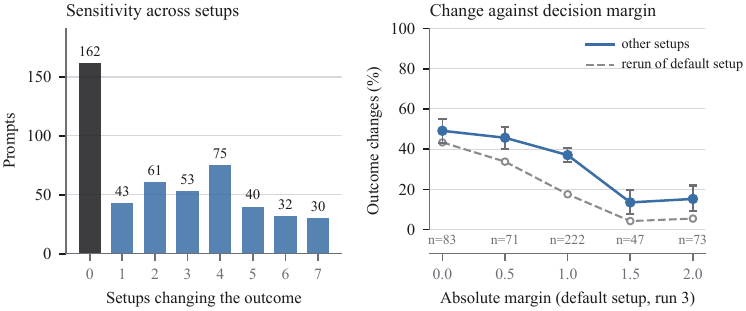}
\caption{Prompt sensitivity across instruction setups and against the default setup's decision margin. Left: how many of the seven other instruction setups change each prompt's outcome relative to the default setup, over the 496 prompts with a verdict in every run; comparing first runs, 162 prompts are unchanged throughout (131 when every run is compared). Right: mean outcome-change rate at each absolute margin, averaged over the seven setups, with whole-prompt bootstrap intervals, against the same rate for an unchanged rerun of the default setup. The margin is taken from a third run of the default setup that is not used to count changes. Point labels give the number of prompts at each margin. Both rates fall as the margin grows and level off at the two highest margins. The instruction setups change more prompts than the rerun at every margin, and the difference is smallest at a margin of zero.}
\label{fig:sensitivity-margin}
\end{figure}

\FloatBarrier

\section{Additional robustness checks}
\label{app:robustness}

The excess over test--retest disagreement varies with the margin. Comparing the mean rate across the seven other instruction setups with test--retest disagreement on the same prompts, taken as the average of the default setup's rerun rate and the setups' own rerun rates, gives an excess of $+18.8$ points at a margin of zero, then $+21.5$, $+23.4$, $+7.6$ and $+10.3$ at the four higher margins, all with intervals above zero. The excess is largest at intermediate margins. Appendix~\ref{app:figures} shows the rates by margin.

Restricting the scope analysis to coding prompts produces a 20.7\% boundary-change rate, against 26.1\% for mathematics prompts and 23.3\% for the full sample. Read against test--retest disagreement on the same prompts, the coding-only excess is $+2.2$ points (95\% CI $-0.4$ to $+5.0$) and the mathematics excess is $+5.5$ points ($+2.5$ to $+8.8$). Where the coding-assistance description is accurate, its effect is indistinguishable from rerunning the default setup; the excess is larger for mathematics prompts, for which the description is inaccurate. The full-sample estimate therefore combines task framing with inaccurate framing for mathematics prompts and is not a clean effect of supplying correct domain information.

Counting a score of $\pm 0.5$ as a tie rather than a win lowers the default setup's test--retest disagreement from 21.2\% to 19.3\% and leaves every excess above zero.

Using paired complete cases prevents missing Gemini outputs from changing any comparison through sample composition. The 15 prompts lost to Gemini parse failures do not differ from the retained prompts in relative response length or markdown density, and are spread across both task categories and all three outcomes of the default setup.

In the May runs, across ten comparisons involving relative response length and markdown density, one test is nominally significant but none remains significant after Holm correction. This does not rule out style as a mechanism. Applying Arena-Hard's style features with coefficients held fixed at the default setup's values subtracts the same quantity from both sides of a paired comparison, so it leaves every value in Table~\ref{tab:main} unchanged and tests nothing. In the May runs, refitting the coefficients within each condition instead gave adjusted differences from the default setup of $-0.15$ points for Gemini, $-0.56$ for the criteria package, $+1.44$ for coding-assistance framing, $-0.74$ for label-only output and $-1.95$ for Claude. That fit absorbs each judge's own response to length and formatting into its coefficients, so it does not isolate a style effect either. We therefore report the raw paired changes and do not claim that style is excluded.

A third judge substitution, Gemini 2.5 Pro, was started and stopped after 129 prompts when it exceeded the run's cost and rate limits. On the 123 of those prompts where Gemini 2.5 Flash also returned a complete outcome, Pro moved the paired aggregate by $-2.34$ points with a 38.2\% boundary-change rate, against Flash's $+0.10$ points and 41.5\%. We report this incomplete run because the completed substitutions were not selected on their results.

Most Arena-Hard prompts have no verifiable answer, but a few do, and one of them is worth reporting on its own. A competitive-programming prompt supplies three worked examples in its statement, so the submissions can be checked against the supplied examples by compiling and running each one rather than by consulting a judge. The candidate model's submission fails all three examples; the baseline model's passes all three. In the May runs, the default setup nonetheless recorded a decisive win for the failing submission. The accuracy-prioritising criteria, both judge substitutions and three of the multi-contrast conditions preferred the submission that passed all three examples, several at the maximum verdict, and the August repeat of the default setup reversed the verdict. We report this as a single case rather than a rate. It shows that a difference in setup can decide whether a verifiably incorrect answer wins, not how often that happens.

In three further examples, both responses passed the checks we applied. On two of them we executed both submissions and each reproduced the expected output on every example supplied in the prompt; on the third, a trigonometric equation, both models report the same correct value. The two responses differ in presentation rather than in content. The judge nonetheless named a winner on each, and in the May runs the winner was not stable: each of the three received a candidate win under at least one condition, a baseline win under at least one other, and a tie under the remainder. On two of the three, every non-tie score had magnitude 0.5, so the recorded wins arise from a tie in one answer order and a slight lean in the other. Correctness alone does not distinguish these responses, yet the sign rule classifies these half-point scores as wins.

\section{Judge instruction packages}
\label{app:prompts}

Crossing the two criteria packages with the two output formats gives the four system prompts below. The scope preamble is prepended to the user message, so it is orthogonal to the system prompt and gives the eight instruction setups. All four packages use the same five verdict labels.

\subsection{Arena-Hard criteria, reasoning required (default setup)}
\label{app:pkg1}

\begin{quote}\small
Please act as an impartial judge and evaluate the quality of the responses provided by two AI assistants to the user prompt displayed below. You will be given assistant A's answer and assistant B's answer. Your job is to evaluate which assistant's answer is better.

Begin your evaluation by generating your own answer to the prompt. You must provide your answers before judging any answers.

When evaluating the assistants' answers, compare both assistants' answers with your answer. You must identify and correct any mistakes or inaccurate information.

Then consider if the assistant's answers are helpful, relevant, and concise. Helpful means the answer correctly responds to the prompt or follows the instructions. Note when user prompt has any ambiguity or more than one interpretation, it is more helpful and appropriate to ask for clarifications or more information from the user than providing an answer based on assumptions. Relevant means all parts of the response closely connect or are appropriate to what is being asked. Concise means the response is clear and not verbose or excessive.

Then consider the creativity and novelty of the assistant's answers when needed. Finally, identify any missing important information in the assistants' answers that would be beneficial to include when responding to the user prompt.

After providing your explanation, you must output only one of the following choices as your final verdict with a label:

1. Assistant A is significantly better: \texttt{[[A>{}>B]]}\\
2. Assistant A is slightly better: \texttt{[[A>B]]}\\
3. Tie, relatively the same: \texttt{[[A=B]]}\\
4. Assistant B is slightly better: \texttt{[[B>A]]}\\
5. Assistant B is significantly better: \texttt{[[B>{}>A]]}

Example output: ``My final verdict is tie: \texttt{[[A=B]]}''.
\end{quote}

\subsection{Arena-Hard criteria, label only}
\label{app:pkg2}

\begin{quote}\small
Please act as an impartial judge and evaluate the quality of the responses provided by two AI assistants to the user prompt displayed below. You will be given assistant A's answer and assistant B's answer. Your job is to evaluate which assistant's answer is better.

Consider whether each assistant's answer is helpful, relevant, and concise. Helpful means the answer correctly responds to the prompt or follows the instructions. Relevant means all parts of the response closely connect or are appropriate to what is being asked. Concise means the response is clear and not verbose or excessive. Also consider creativity and novelty when relevant, and identify any missing important information.

Output only one of the following choices as your final verdict with a label, with no explanation:

1. Assistant A is significantly better: \texttt{[[A>{}>B]]}\\
2. Assistant A is slightly better: \texttt{[[A>B]]}\\
3. Tie, relatively the same: \texttt{[[A=B]]}\\
4. Assistant B is slightly better: \texttt{[[B>A]]}\\
5. Assistant B is significantly better: \texttt{[[B>{}>A]]}

Example output: ``\texttt{[[A=B]]}''.
\end{quote}

Relative to~\ref{app:pkg1} this removes the self-generation step, the comparison against the judge's own answer, the error-correction instruction and the clarification-on-ambiguity sentence, in addition to the explanation itself. This is why we treat the output contrast as a package.

\subsection{Accuracy-prioritising criteria, reasoning required}
\label{app:pkg3}

\begin{quote}\small
Please act as an impartial judge and evaluate the quality of the responses provided by two AI assistants to the user prompt displayed below. You will be given assistant A's answer and assistant B's answer.

Evaluate the responses on three criteria with explicit priority:

CORRECTNESS (primary): Are the technical claims, code, calculations, and reasoning accurate? Is the answer factually right?

HELPFULNESS (constraint): Does the answer address the user's actual request? Is it well-organised and easy to follow?

SAFETY (constraint): Does the answer avoid harmful, deceptive, or unethical content?

In your final verdict, prioritise correctness. A response that is more helpful but less correct should be rated lower than a response that is correct but less helpful. Helpfulness and safety are constraints, not primary criteria.

Begin by analysing each response against the three criteria, then state your final verdict.

Output one of the following as your final verdict with a label:

1. Assistant A is significantly better: \texttt{[[A>{}>B]]}\\
2. Assistant A is slightly better: \texttt{[[A>B]]}\\
3. Tie, relatively the same: \texttt{[[A=B]]}\\
4. Assistant B is slightly better: \texttt{[[B>A]]}\\
5. Assistant B is significantly better: \texttt{[[B>{}>A]]}

Example output: ``My final verdict is: \texttt{[[A=B]]}''.
\end{quote}

\subsection{Accuracy-prioritising criteria, label only}
\label{app:pkg4}

\begin{quote}\small
Identical to~\ref{app:pkg3} through the priority statement, then: ``Output only one of the following as your final verdict with a label, with no explanation'', followed by the same five labels and the example output ``\texttt{[[A=B]]}''. The sentence about helpfulness and safety being constraints rather than primary criteria is also dropped.
\end{quote}

\subsection{Scope preamble and case template}
\label{app:pkg5}

The coding-assistance setups prepend the following to the user message:

\begin{quote}\small
The following user prompt is from a coding-assistance task. Evaluate the responses with that context in mind.
\end{quote}

The case itself is supplied in Arena-Hard's own template, with the user prompt and both answers delimited by \texttt{<|User Prompt|>}, \texttt{<|The Start of Assistant A's Answer|>} and the corresponding end markers.

\section{Conditions and counts}
\label{app:conditions}

Table~\ref{tab:conditions} lists the eight GPT-4.1 setups from the October runs. A prompt is complete when both answer orders returned a parsed verdict in every run of that setup. The paired column counts prompts complete in every run of both the setup and the default setup. The tie rate is the proportion of complete prompts whose order-averaged score is exactly zero, averaged over runs. In the May runs, Claude and Gemini returned 498 and 485 complete prompts, 494 and 481 of them paired with the default setup.

\begin{table}[!htbp]
\centering
\caption{The eight GPT-4.1 setups in the October runs, with counts, raw scores by run and tie rates.}
\label{tab:conditions}
\includegraphics[width=\linewidth]{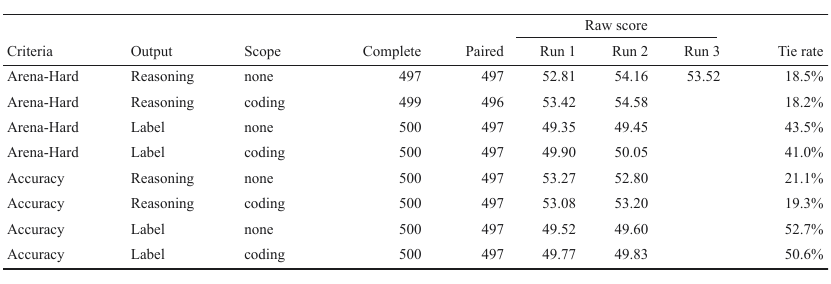}
\end{table}

The default setup was run three times on the same day, scoring 52.81, 54.16 and 53.52. Between pairs of runs the paired score change is $+1.21$ (95\% CI $+0.05$ to $+2.39$), $+0.68$ ($-0.50$ to $+1.86$) and $-0.53$ ($-1.71$ to $+0.65$), and 21.1\%, 21.5\% and 20.9\% of prompts change winner. Linearly weighted $\kappa$ between runs averages 0.763 (95\% CI 0.730 to 0.794). In the May runs, the August repeat moved the score by $+0.10$ points and changed 20.0\% of winners. Every raw score in the table is computed on that setup's own complete prompts, so the paired figures differ slightly from the tabulated ones.

In May, the ten original conditions were issued as 10,000 calls, of which 9,996 were retained: two Claude prompts errored in both answer orders and were not recovered. Of the 9,996 retained, 24 returned no parsable verdict. Eighteen of these were in the Gemini condition, where the failures were mixed: eight reached the 16,000-token cap, four were degenerate generations that emitted several hundred thousand characters of trailing whitespace, and six were ordinary-length outputs that simply never produced a bracketed label. The remaining six failures were five in the default setup and one in the coding-scope condition, all of which did reach the token cap. The other six instruction conditions had complete parse coverage. The default setup's losses are why 496 rather than 500 prompts propagate through every paired comparison in the May runs. GPT-4.1 and Gemini used a 16,000-token output cap; Claude used 4,096, which one of 996 Claude rounds reached exactly. Counting the repeat, the abandoned Gemini 2.5 Pro run and the Claude retries, roughly 11,300 calls were issued in total. The October runs issued a further 17,000 calls, of which 16,996 returned a parsed verdict.

\section{Verdict parsing}
\label{app:parsing}

Calls that raised a rate-limit error were retried automatically, but an output that returned without a parsable verdict was not re-issued in the GPT-4.1 conditions, and was treated as missing. Two exceptions: fourteen Claude prompts whose verdicts failed to parse were re-run several days later, and the Gemini runner re-queued parse failures on resume.

Verdicts are extracted with a primary expression matching the double-bracketed label, and a fallback expression matching a single-bracketed label. Both search the uppercased output and take the final match, so a judge that discusses the labels before committing is scored on its final statement. An output is invalid when neither expression matches, and invalid outputs are never imputed. Because a prompt requires both answer orders to be valid, one invalid output removes that prompt from every comparison involving that condition.

\section{Estimators}
\label{app:definitions}

For prompt $i$, condition $c$ and answer order $o$, let $r_{i,c,o}\in\{-2,-1,0,1,2\}$ be the verdict after recoding, so that positive values favour the candidate in both orders. The order-averaged score is
\begin{equation}
s_{i,c}=\tfrac{1}{2}\left(r_{i,c,1}+r_{i,c,2}\right),
\end{equation}
which takes half-integer values from $-2$ to $2$, and is undefined if either order failed to parse. Its sign class is $z_{i,c}=\operatorname{sign}(s_{i,c})\in\{-1,0,1\}$, representing a baseline win, a tie and a candidate win.

The raw win-equivalent aggregate score over the $n_c$ complete prompts is
\begin{equation}
\widehat{A}_c
=\frac{100}{n_c}\sum_i\frac{s_{i,c}+2}{4}
=25\,\bar{s}_c+50 .
\end{equation}

Let $\mathcal{P}_{c,c'}$ be the prompts complete under both conditions. The boundary-change rate and paired aggregate change are
\begin{equation}
\begin{aligned}
\widehat{F}(c,c')
  &=\frac{1}{|\mathcal{P}_{c,c'}|}
    \sum_{i\in\mathcal{P}_{c,c'}}
    \mathbb{1}\!\left[z_{i,c}\neq z_{i,c'}\right],\\
\widehat{\Delta}_{c-c'}
  &=\frac{25}{|\mathcal{P}_{c,c'}|}
    \sum_{i\in\mathcal{P}_{c,c'}}
    \left(s_{i,c}-s_{i,c'}\right).
\end{aligned}
\end{equation}

A direct reversal is the stricter event $z_{i,c}\,z_{i,c'}=-1$, which excludes movement into or out of a tie but keeps tied prompts in the denominator. Every interval is a percentile bootstrap interval over prompts.

For a factor $X$, let $\mathbf{z}$ index the four combinations of the other two factors, over the 496 prompts complete in every run. The marginal aggregate effect and marginal boundary-change rate are
\begin{equation}
\begin{aligned}
\widehat{\delta}_X
  &=\frac{1}{496}\sum_i\frac{25}{4}\sum_{\mathbf{z}}
    \left(s_{i,X=1,\mathbf{z}}
    -s_{i,X=0,\mathbf{z}}\right),\\
\widehat{B}_X
  &=\frac{1}{496}\sum_i\frac{1}{4}\sum_{\mathbf{z}}
    \mathbb{1}\!\left[
      z_{i,X=1,\mathbf{z}}
      \neq z_{i,X=0,\mathbf{z}}
    \right].
\end{aligned}
\end{equation}

Here each setup's score is averaged over its runs, and the boundary-change rate is averaged over every pairing of a run of one setup with a run of the other.

Each setup $c$ has runs $c^{(1)},\dots,c^{(k_c)}$: three for the default setup $c_0$ and two for every other setup. Let $\widehat{F}(c,c')$ now denote the boundary-change rate averaged over every pairing of a run of $c$ with a run of $c'$, and let the test--retest disagreement $\widehat{F}_c$ be the rate averaged over pairs of runs of $c$. The excess over both setups' test--retest disagreement is
\begin{equation}
\widehat{E}(c)=\widehat{F}(c,c_0)-\tfrac{1}{2}\bigl(\widehat{F}_c+\widehat{F}_{c_0}\bigr),
\end{equation}
and the excess against the default setup's test--retest disagreement alone, the only version available in May, is $\widehat{F}(c,c_0)-\widehat{F}_{c_0}$. All rates are computed on the prompts complete in every run involved, and each bootstrap resample of prompts recomputes every term, so the difference is properly paired.

The sensitivity of a prompt is the number of the seven other instruction setups in which its sign class differs from the default setup's, comparing first runs, and its margin is $|s_{i,c_0}|$ taken from the default setup's third run.

Before computing linearly weighted Cohen's $\kappa$, order-averaged scores are re-binned to five tiers at cut points $-1.5$, $-0.5$, $0.5$ and $1.5$, with the central tier closed at both ends. A score of exactly $\pm0.5$ therefore counts as a tie for $\kappa$ while counting as a win for the sign class, so $\kappa$ is slightly more conservative than the boundary-change measure.

\section{Threshold-based classification}
\label{app:thresholds}

As a descriptive summary we also classify each prompt by how consistent its sign class is across the eight instruction setups. A prompt is \emph{unanimous} when every condition assigns the same sign class. Otherwise it is \emph{majority} when one sign class holds at least a majority share and the tie share is not too large, and \emph{unresolved} when neither holds. Using every October run as a condition, at the default thresholds of a 0.6 majority share and a 0.4 tie share, 132 prompts (26.4\%) are unanimous, 136 (27.2\%) are majority and 232 (46.4\%) are unresolved, on all 500 prompts.

Table~\ref{tab:thresholds} sweeps both thresholds. The unanimous share is 26.4\% at every grid point, because unanimity is evaluated before either threshold is consulted; it is a property of the definition and not evidence of robustness. The unresolved share moves from 33.6\% to 54.4\%, which is why we do not treat this partition as a natural division of the prompts and report the threshold-free unanimity result in the main text instead.

\begin{table}[!htbp]
\centering
\caption{Threshold sweep over all 500 prompts, using run 1 of each setup and all 17 October runs. The unanimous share is invariant by construction.}
\label{tab:thresholds}
\includegraphics[width=\linewidth]{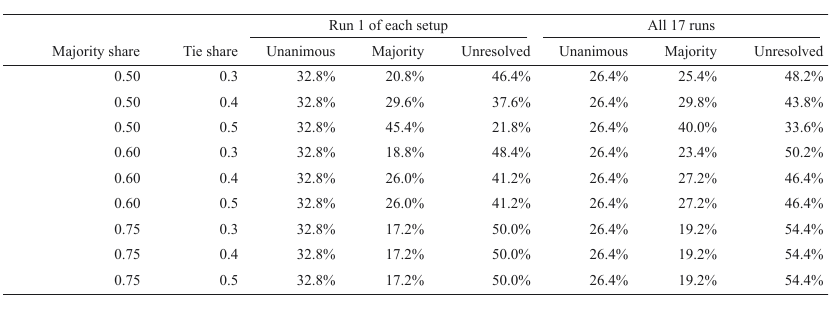}
\end{table}

The main text reports 131 of 496 (26.4\%) rather than 132 of 500 (26.4\%). The two differ because the main text restricts to prompts complete in every run, whereas this appendix follows the classification code in dropping missing runs and classifying on those that remain. One of the four prompts with an incomplete run is unanimous across the runs it does have, which is why the count rises by one as well as the denominator by four.

Restricting the classification to the four GPT-4.1 conditions that require reasoning together with the Claude substitution leaves 17.6\% unresolved in the May runs. A set of five conditions is mechanically easier to satisfy than a set of eight, so this figure should be compared against equally sized subsets rather than against the eight-condition result.

\FloatBarrier

\section{Possible implications for uses of individual verdicts}
\label{app:exploratory}

Nothing in this appendix was tested. Each entry reasons from a measured mechanism, that 21.2\% of prompt-level outcomes change on a rerun which changed nothing, to an unmeasured consequence in some setting where LLM judges are used. It is a set of hypotheses about where that instability would matter, each of which would need its own experiment. We include it because the practical question of who should care is separable from the measurement, and because several of these settings are cheap to check.

The measurement behind it is also narrow. One benchmark, one model pair chosen because its published comparison sits near 50\%, one judge family, and repeats run on a single day on one pinned model version. We expect a model pair with a wide capability gap to show lower instability, since the prompts that never change here are those where one model plainly wins.

\begin{table}[!t]
\caption{Settings where per-item judge outputs are consumed, and what the measured instability would and would not disturb in each. Hypotheses, not results.}
\label{tab:exploratory}
\centering
\small
\renewcommand{\arraystretch}{1.25}
\begin{tabular}{@{}>{\raggedright\arraybackslash}p{0.19\linewidth}>{\raggedright\arraybackslash}p{0.43\linewidth}>{\raggedright\arraybackslash}p{0.32\linewidth}@{}}
\toprule
Setting & What would be disturbed & What would survive \\
\midrule

Preference data for reinforcement learning
& Reward models train on per-item labels, which is the unstable layer. Where both responses are correct, the label asserts that one correct answer is worse than another, which may train presentation rather than quality.
& Comparisons with a decisive margin, where a real quality difference exists to learn. \\

Safety review
& Flagging a response is a per-item verdict, and the contested cases that most need to be right may be the ones that move. Measured here on pairwise comparison rather than on classification, so a classification judge's rate would need measuring separately.
& Cases on which every configuration we tested agrees, if these are the clear-cut ones. \\

Model selection
& Rerunning the default setup moves the aggregate by 0.5 to 1.2 points, but changing the judge moves it by up to four. For this model pair, ranking is largely robust to run-to-run variation, not to judge choice.
& Comparisons where the gap exceeds a few points. \\

Regression testing
& A set of prompts that newly fail cannot be separated from run-to-run movement without a repeat.
& Large regressions on decisive items. \\

Audit and regulation
& A rule citing a benchmark inherits its aggregation choices. An aggregate score is stable; a list of failed prompts is not, so a per-item audit would need a repeat protocol written into it.
& Comparative claims between systems, reported with intervals. \\

Published comparisons
& A three-point aggregate gap sits inside the range a judge change produces here, and a claim about the proportion of prompts won is less stable still.
& A large, decisive gap. \\

Contractual thresholds
& A per-item pass rate is a threshold on the unstable layer, and the rule converting verdicts to passes is itself a design choice that is rarely disclosed.
& Little, without a stated repeat protocol. \\

\bottomrule
\end{tabular}
\end{table}

\end{document}